\documentclass[conference]{IEEEtran}
\IEEEoverridecommandlockouts
\usepackage{cite}
\usepackage{amsmath}
\usepackage{amsmath,amssymb,amsfonts}
\usepackage{algorithmic}
\usepackage{graphicx}
\usepackage{textcomp}
\usepackage{xcolor}
\usepackage{hyperref}

\def\BibTeX{{\rm B\kern-.05em{\sc i\kern-.025em b}\kern-.08em
    T\kern-.1667em\lower.7ex\hbox{E}\kern-.125emX}}

\usepackage{eso-pic}
\newcommand\AtPageUpperMyright[1]{\AtPageUpperLeft{
 \put(\LenToUnit{0.5\paperwidth},\LenToUnit{-1cm}){
     \parbox{0.5\textwidth}{\raggedleft\fontsize{9}{11}\selectfont #1}}
 }}
\newcommand{\conf}[1]{
\AddToShipoutPictureBG*{
\AtPageUpperMyright{#1}
}}

\begin{document}

\title{Automated Intersection Management with MiniZinc}
\conf{2020 2nd International Conference on Sustainable Technologies for Industry 4.0 (STI), 19-20 December, Dhaka}

\author{
    \IEEEauthorblockN{
    Md. Mushfiqur Rahman\IEEEauthorrefmark{1},
    Nahian Muhtasim Zahin\IEEEauthorrefmark{2},
    Kazi Raiyan Mahmud\IEEEauthorrefmark{3}, 
    Md. Azmaeen Bin Ansar\IEEEauthorrefmark{4}}
    \IEEEauthorblockA{
    Islamic University of Technology, Gazipur
    \\\{ 
    mushfiqur11\IEEEauthorrefmark{1}, 
    nahianmuhtasim\IEEEauthorrefmark{2},
    kaziraiyan\IEEEauthorrefmark{3},
    azmaeen\IEEEauthorrefmark{4}\}
    @iut-dhaka.edu}
}

\IEEEoverridecommandlockouts
\IEEEpubid{\makebox[\columnwidth]{978-1-7281-9576-6/20/\$31.00~\copyright2020 IEEE \hfill} \hspace{\columnsep}\makebox[\columnwidth]{ }}

\maketitle

\begin{abstract}
Ill-managed intersections are the primary reasons behind the increasing traffic problem in urban areas, leading to nonoptimal traffic-flow and unnecessary deadlocks. In this paper, we propose an automated intersection management system that extracts data from a well-defined grid of sensors and optimizes traffic flow by controlling traffic signals. The data extraction mechanism is independent of the optimization algorithm and this paper primarily emphasizes on the later one. We have used MiniZinc \cite{nethercote2007minizinc} modeling language to define our system as a constraint satisfaction problem which can be solved using any off-the-shelf solver. The proposed system performs much better than the systems currently in-use. Our system reduces mean waiting time and standard deviation of waiting time of vehicles, and avoids deadlocks.
\end{abstract}


\section{Introduction}
Road intersections are one of the core parts of city planning and management. Intersections are nodes in traffic network where several roads meet, so managing this node is crucial for efficient traffic flow. In developed countries these intersections are managed by strict traffic rules and regulations. Developing countries have a different scenario since drivers are more reckless and traffic police is used to maintain order in the road. These intersection points are the most important factors for a city’s overall productivity and efficiency. By reducing traffic congestion occurring at these intersection points, productivity can be improved. In this paper our aim is to reduce the average delay that occurs in intersection nodes. Traditionally in Bangladesh, a congestion based traffic management system is used. That is, lanes are opened based on the congestion of vehicles in that lane. This is not an optimum system and this results in lots of deadlocks at intersections every day. Poor traffic system results in huge wastage of man-hour every day. As a result, overall productivity of the city decreases \cite{harriet2013assessment}. Traffic congestion not only causes millions of dollars worth loss every year but also worsens the city's environment by aggravating air pollution. It is of utmost importance to have a foolproof traffic management system in order to create a sustainable and growing city. An organized traffic management system will not only reduce the average delay of cars but also alleviate pressure from roads, and reduce energy consumption, and carbon emission of vehicles.

We propose a model based on MiniZinc that takes two factors into account to control the traffic flow: vehicle priority and vehicle waiting time. Our intuition is, traffic management can be dealt as an optimization problem where the goal would be to reduce some cost function that minimizes traffic delay while ensuring priority for emergency vehicles.

There have been several traffic management system overtime. Various scheduling and queuing models have been used to find the optimum way. In order to optimize the flow of vehicles and minimize the average delay it is of utmost importance to know the current condition of the road. An array based system is formulated to simulate 12 lanes of an intersection and that array is updated based on the cars that arrive after each iteration. 2nd part consists of optimization of traffic flow. Here waiting time and priority of vehicle is taken as factors to minimize the average delay. But since calculating average time for all the possible combination in a 12 lanes will create a huge overhead we decided to use heuristics to reduce the search space and eliminate erroneous nodes. Minizinc was used to create our model and minimize the average time. We took several constraints into account in our model to devise as much realism as possible.

\begin{figure}
    \centering
    \includegraphics[width=0.9\linewidth]{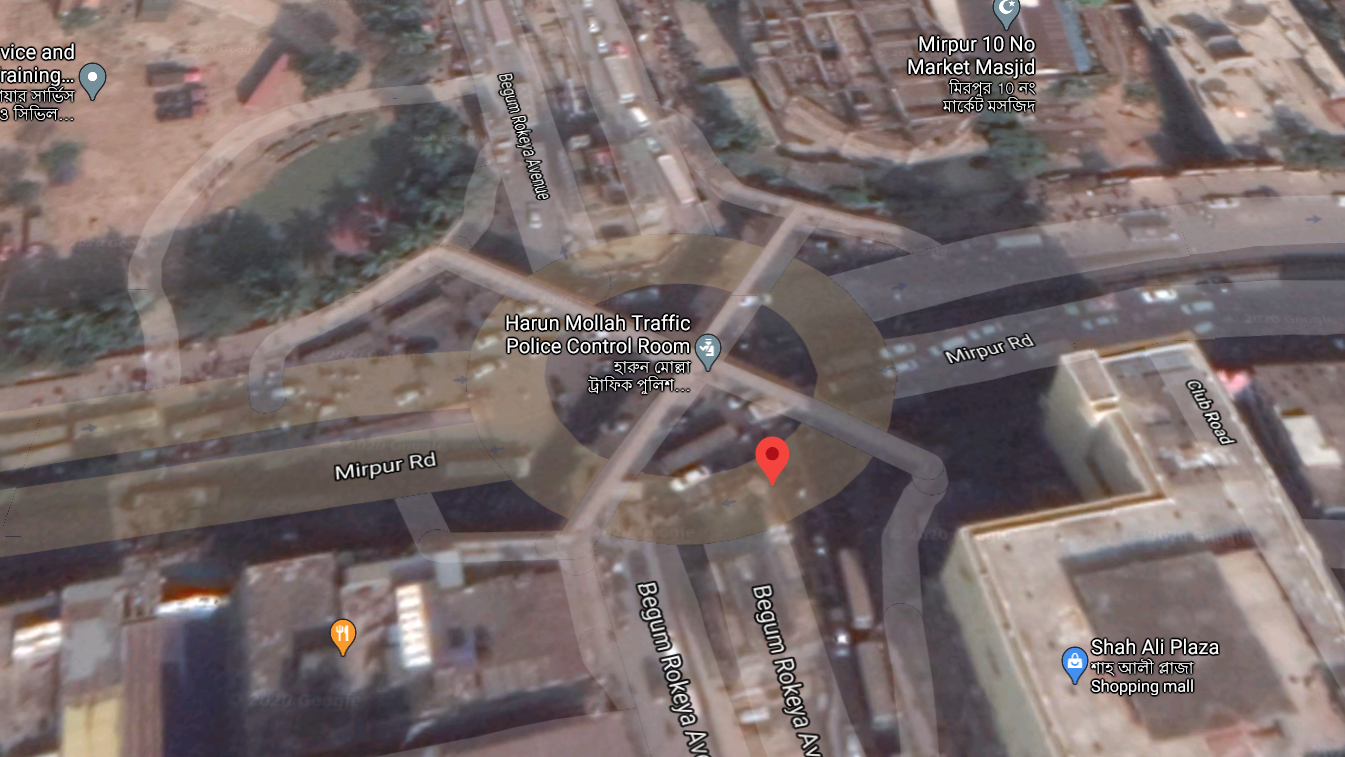}
    \caption{Satellite view of an intersection merging four roads \cite{satelliteview}}
    \label{satelliteview}
\end{figure}

\section{Related Works}

An innumerable number of intersection traffic control mechanisms and procedures exist each with their same set of goals and with different types of agents. Some of them use different methods of scheduling and queuing some examples include a method \cite{ahmad2014earliest} where the earliest-deadline-based scheduling is used to reduce congestion and ultimately reduce delay times and another method \cite{dresner2008multiagent} where the FCFS (First Come First Serve) protocol is used to control traffic flow.
Most works \cite{dresner2008multiagent,chen2015cooperative,perronnet2019deadlock,katriniok2017distributed,jin2012advanced,lee2013sustainability} that are pertinent to intersection traffic management have been based on autonomous vehicles. The objective of these works have been different, some \cite{jin2012advanced,lee2013sustainability} dealing with the reduction of emissions while others have criteria such as goals such as congestion avoidance. The goal of the method proposed in this article is to reduce delay of the vehicle i.e. the time a vehicle waits motionless. Some methods employ neural network and data mining techniques \cite{parker2017best,labib2019integrating}, whilst others are strictly rule based \cite{slavin2013statistical}. The methods which are most relevant are discussed.

When looking at contemporary intelligent traffic control mechanisms, the most common is the SCATS \cite{slavin2013statistical, wikipedia} (Sydney Cooperative Adaptive Traffic System). SCATS implements traffic control by using sensor data which is based on the number of vehicles and pedestrians to set the optimal phasing i.e. cycle time, phase, offsets and splits for a given traffic scenario. It has LOCAL and MASTER layers,  the LOCAL layer determines the local traffic snapshot of an area through sensors and the MASTER layer is a remote controller which mandates the phasing. Although SCATS has a three level priority system, it is only reserved for public transport timings such as busses and trams. Priority assignment to all vehicles in the system in SCATS is not done. Also SCATS gives precedence to the lane which is the most congested. Compared to the SCATS, the method outlined in this article, assigns a priority to each vehicle and also considers the waiting time based on intermittent snapshots taken of the scenario. Another prominent method \cite{chen2015cooperative}, uses the discretization of the intersection, i.e. segregates a four way intersection into discrete blocks and frames the problem as a discrete optimization problem of resource allocation where the resource is each individual block. The paper describes a system where individual autonomous agents reserve the optimal set of blocks by communicating among themselves based on their predetermined path. Such a method is not feasible in the context of Bangladesh due to the general unavailability of autonomous vehicles and also the road scenarios being vastly random for discretization as apart from cars there may exist a lot of different agents. A  centralized model  \cite{dresner2008multiagent} version, where the vehicular agents communicate with a centralized traffic controller to reserve spaces with reference to the destined path will present similar nuisances. Another method of note uses multitude of sensors in tandem with autonomous vehicles to control traffic at intersections. The system uses FCFS queuing to control the movement of vehicles. The problem associated with FCFS is that perfect traffic flow conditions are assumed and it does not account for unexpected pedestrian activity.

Collecting traffic data is not an easy process especially when there is not a centralized traffic system that tracks and updates the real-time traffic data in regular intervals. A manual approach would be cumbersome as the data collector needs to count each car manually with a decent accuracy. Another major problem of collecting traffic data is that the data varies significantly depending on a lot of other factors like time of the day, special occasions etc. Like our proposed system, any other traffic system that aims to reduce traffic delay needs to have access to clean and accurate data for performing efficiently. Several innovative approaches to collect traffic data using GPS data \cite{rahman2018traffic}, real-time traffic data from Google Map \cite{zafri2020effectiveness}, gravity model \cite{sayed2017understanding}, artificial neural network \cite{siddiquee2017predicting} were proposed which showed some excellent results.

\section{Methodology}
This paper proposes a two step process to solve the problem of intersection management. In the first step, the system generates the traffic flow data at a particular instance. In the second step, the system uses heuristic method to optimize the traffic flow by controlling the traffic signals at intersections. This paper primarily focuses on the later step.

\subsection{Traffic Flow Generation}
To optimize the flow of vehicles at an intersection and to minimize the average delay of the vehicles, the first task is to know the current condition of the traffic flow at that intersection. In this paper, a detailed elaboration is not given for this task but rather an outline for the task has been laid out.

To be specific for our system, a sensor system laid out in a specific manner generates a 3D array of traffic flow. For each car passing the sensor, a new entry is added to the array.
\subsubsection{Sensor-Layout}
For generalization, our implementation considers 12 possible commute routes for a four-way intersection. In this paper, these commute routes are considered as paths. To obtain the number of cars in each of the paths, some hardware system has to be in place. Salama, Saleh and Eassa \cite{5646059} proposes an elaborate system of sensor-layout for the task that we are targeting. They have used photo-electric sensors prior to and after traffic lights. In our implementation, we have adopted their sensor-layout and have not made any significant changes to it. We program the sensors in such a way that they generate arrays with specifications required for our AI system.

\subsubsection{Array Specification}
As we consider 12 paths for our system, the generated array also consists of 12 tuples -- one for each path. Each tuple stores two queues -- priority queue and waited time queue. 

The priority queue is, in fact, a list of cars in the system in that particular path. Each car is associated with a priority. The sensors pick this priority by determining the car. Emergency vehicles like ambulances, fire trucks etc. have the highest priority. Public transports that can carry large number of people, for example, busses, also have a high priority. The remaining slots in the list are marked with zeros to indicate no priority or no car.

For each vehicle in the queue, along with the priority, a waiting time is also used. Once a sensor identifies a car and registers it in the priority queue, it also starts a counter. This counter stores the time this vehicle stays in the system.

Since the entire data structure is in the form of an array, the queues are limited to a maximum length. For a given intersection, it is also quite intuitive to have a maximum length of cars since the size of the road leading to the intersection is limited.

Therefore, the dimension of the array becomes \textit{(number of paths, 2, max queue length)}. 

\subsubsection{Array Update}
The set of sensors update the array when new vehicles joins the system or vehicles in the system exits the queue. Another major update is the counter of the waiting time for the vehicles already in the queue. This update is calculated by the mother system.

\subsection{Traffic Flow Optimization}
In our implementation, the system searches for the optimum output from a vast pool of outputs. The optimum output is the setting that minimizes the average delay. The system needs to calculate the total waiting time of all the vehicles in the system to calculate the average waiting time. So to minimize the waiting time, the system needs to calculate the average waiting time for all possible options. For a 12 path system, if any number of paths can be open at a time, the number of possible combinations of path becomes $\sum_{n=1}^{12}{12 \choose n}$. If the next $t$ signals are to be generated, the number of possible path becomes 
$$ \left(\sum_{n=1}^{12}{12 \choose n}\right)^{t} $$

Calculating the average time for all these numbers requires a huge overhead and is not feasible in real time. The only way around it is the use of heuristics to reduce the search space and propagate towards the end node while eliminating the erroneous nodes. 

The main contribution of this paper is the creation of an artificially intelligent system that is capable of using the traffic flow data to optimize the movement of vehicles in the streets, to prevent deadlocks and to minimize the average delay of the vehicles in the system. This system was created leveraging the power of the constraint modeling language, MiniZinc \cite{nethercote2007minizinc}. This paper is not the first one to treat intersection management as an optimization problem \cite{christofa2011traffic,sen1997controlled,hausknecht2011autonomous}. The other papers have used various optimization techniques. Our paper creates a series of constraints and optimizes the data flow by satisfying the constraints while minimizing the average delay.

\subsubsection{Environment}
Our proposed system mainly uses the constraint programming language, MiniZinc. The constraints are laid out in MiniZinc and with an off-the-shelf solver, the vehicle queue data is analyzed. The result of the analysis is the traffic signal for each path in the next $n$ successive iterations. This series of signals is supposed to minimize the traffic delay on average while keeping the constraints fulfilled.

The output for most solvers are similar and experiments show that none of them have a significant advantage over the others for our model. Therefore, any off-the-shelf FlatZinc supported solver can be chosen for our model.

\subsubsection{Constraints}
Our system defines a series of constraints for this problem and the effectiveness of these constraints determine our system's performance.
\begin{itemize}
    \begin{figure}[hbt]
    \centering
    \begin{tabular}{|c|c|c|c|}
        \hline
         \includegraphics[width=0.19\linewidth]{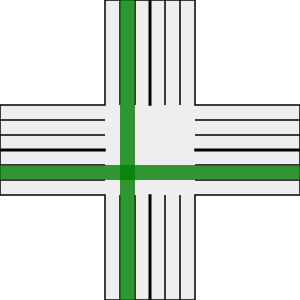}&
         \includegraphics[width=0.19\linewidth]{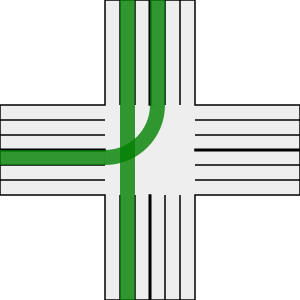}&
         \includegraphics[width=0.19\linewidth]{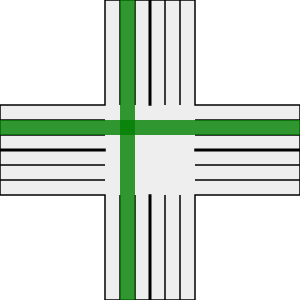}&
         \includegraphics[width=0.19\linewidth]{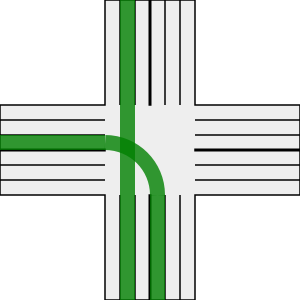}
         \\
         \hline
         \includegraphics[width=0.19\linewidth]{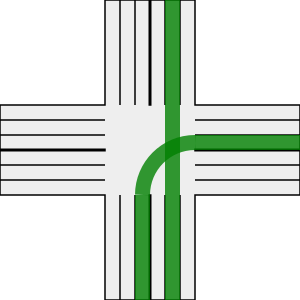}&
         \includegraphics[width=0.19\linewidth]{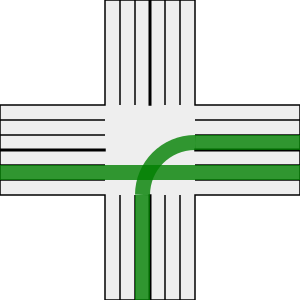}&
         \includegraphics[width=0.19\linewidth]{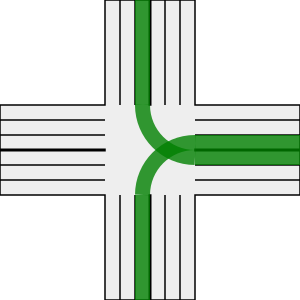}&
         \includegraphics[width=0.19\linewidth]{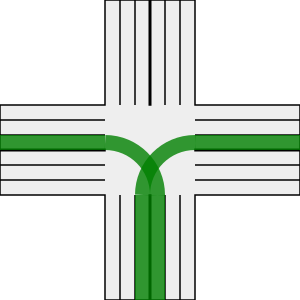}
         \\
         \hline
         \includegraphics[width=0.19\linewidth]{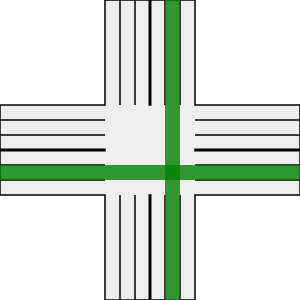}&
         \includegraphics[width=0.19\linewidth]{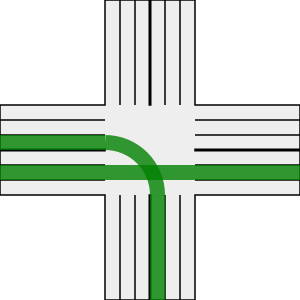}&
         \includegraphics[width=0.19\linewidth]{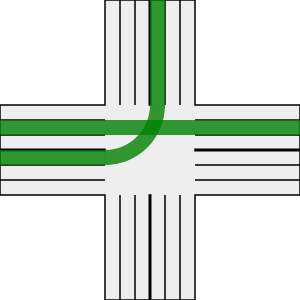}&
         \includegraphics[width=0.19\linewidth]{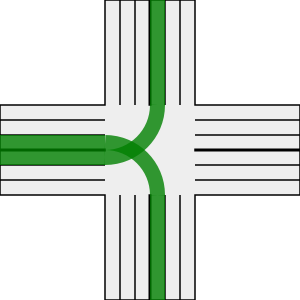}
         \\
         \hline
         \includegraphics[width=0.19\linewidth]{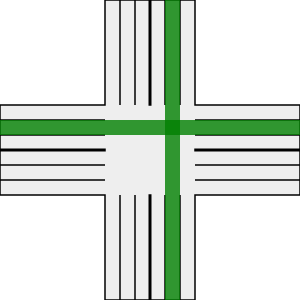}&
         \includegraphics[width=0.19\linewidth]{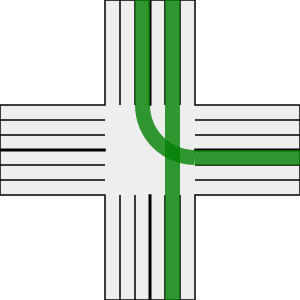}&
         \includegraphics[width=0.19\linewidth]{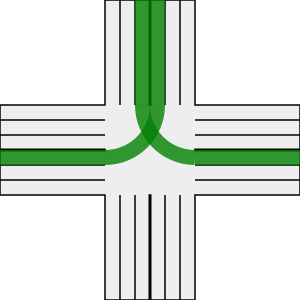}&
         \includegraphics[width=0.19\linewidth]{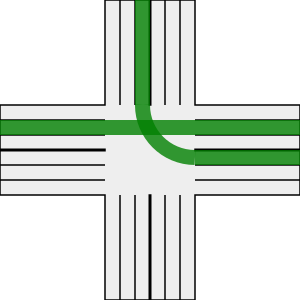}
         \\
         \hline
    \end{tabular}
    \caption{Competing paths}
    \label{fig:constraint}
\end{figure}
    \item \textbf{Competing Path Constraint:} Not all paths in the system intersect each other. But the paths that do intersect, like in Fig.\ref{fig:constraint}, cannot remain open simultaneously. A set of constraint is defined to fulfill this condition. This constraint also helps in reducing the search space.
    \item \textbf{Queue Update Constraint:} The system considers that a fixed time is required for cars to move from one place to another when the number of traffics in the system is on a certain level. Keeping this in mind, a constraint was defined that predicts the future queue condition for certain traffic signal changes.
    \item \textbf{Initial Slow Start Constraint:} When a path is opened from closed, there is an overhead delay due to human nature. This constraint keeps this phenomenon in consideration.
    \item \textbf{Waiting Time Update Constraint:} This constraint calculates the waiting time of the individual vehicles for the chosen path. So, for each vehicle in the system, for a particular set of path remaining open and rest remaining closed, this constraint updates the waiting time.
    \item \textbf{Total Waiting Time Constraint:} This constraint calculates the total waiting time of the system after $k$ successive signal changes. The previously declared constraints are used here. The primary goal of the solver is to generate a list of signals that minimizes the out put of this variable.
\end{itemize}

\subsection{Reaching Solution}
The solution for a given instance is the list of next $k$ subsequent signals in the system that minimizes the total delay. Data obtained from the sensors is passed on to the MiniZinc model and with any solver of choice, the output is obtained.

The implementation assumes a fixed time for propagation of vehicles. On average, this propagation works, and so the predictions are somewhat accurate. But the predictions can go one or two vehicle wrong in every iteration. So, if the original value is not updated after every signal change, the errors can pile up, resulting a massive error. Therefore, though the system calculates for the next $k$ instances, the system again updates its data from the sensors in the immediately next instance. This time, the system generates signal list for $1^{\text{st}}$ to $(k+1)^{\text{th}}$ signal change. So, in spite of generating $k$ subsequent signals, only the immediate one is used every time. Increasing the value of $k$, improves the performance of the model but also massively increases the time required to generate the solutions. On the other hand, decreasing $k$ results in decreased farsightedness of the model.

\section{Results}

The goal of the proposed model is to reduce the delay time endemic to typical intersection traffic. In order to substantiate the reduction in delay brought about by the proposed model, a comparison was made between common traffic control practices in Bangladesh. Model F1 and F2 were chosen for comparison. F1 prioritizes the lanes where there is a higher traffic flow or congestion and allows such lanes to move first and F2 is the traffic control system that allocates an equal time for traffic movement for all lanes in the system.
Due to the inaccessibility of traffic data in official records and also due to the legal implications associated with drone photography, data could not be collected, rather delay and congestion data from \textit{Performance evaluation parameters for signalized road intersections under heterogeneous traffic condition} \cite{kader2016performance} was used. Priority for each individual vehicle was assigned using a random assignment method.There are several factors to consider while dealing with traffic intersection,
e.g. pressure, delay, energy consumption, Carbon Dioxide emission. Average delay was chosen as the main factor as it is the main deciding factor for the overall traffic situation of an area.

\begin{figure}
    \centering
    \includegraphics[width=0.9\linewidth]{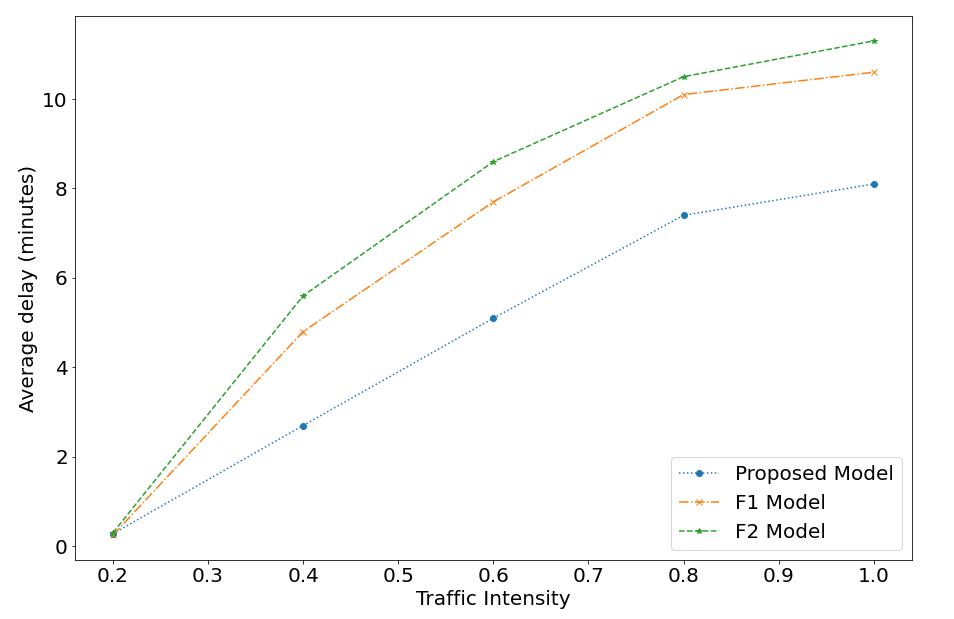}
    \caption{Comparison of performance between existing systems (F1 and F2) and our proposed implementation}
    \label{fig:final_comparison}
\end{figure}

The horizontal axis of the visualization represents the traffic intensity where 1 in the scale signifies a fully packed lane and 0 an empty lane. The vertical axis represents the average delay i.e. the amount of time in minutes on average a vehicle remains stationary in the intersection. From the graph it is evident that the proposed MiniZinc method offers a significant reduction in average delay times for all levels of traffic intensity. F1 and F2 almost hug each other whilst the difference between these two methods and the proposed method grow greater with increasing intensity. At the highest traffic intensity of value 1 a difference of around \textbf{2.5} minutes from F2 and around \textbf{3.2} minutes from F1 is observable.

The proposed MiniZinc model is based on a twelve lane and four way intersection, modelling of a different number of lanes and ways is possible as appropriate.It can also be applied to bigger traffic intersection management scenarios such as Arterial Network or Grid Network.

\section{Discussion}
As evident from the results above, the proposed method significantly performs better in lowering delay times than other commonly practiced methods of traffic management in intersections i.e. a congestion based method and one where the time for each lane is constant. The two phased structure of the method consists of first gathering a general picture of the traffic conditions in each unit lane of the system, which enables the system to be very dynamic and make decisions which is based on the most current scenario of traffic in the system. The level of granularity of the time interval in the algorithm can be adjusted as appropriate. This makes the proposed method much more robust in setting the optimal signal times.

Another important aspect in the algorithm’s decision making process is the priority of the individual vehicles. The priority is important due to the different types of vehicles available in the road traffic system and their importance. An example of this would be an ambulance having higher priority due to the necessity of it reaching its destination being of greater urgency compared to other vehicles for obvious reasons. Other high priority vehicles may include law enforcement and government officials.

The algorithm for the model was constructed using MiniZinc as stated earlier. The high level abstraction provided by MiniZinc allowed for simpler descriptions of all the constraints and variables associated with the model.
A general discussion is necessary of the practical aspects of implementing the model. As explained previously the algorithm works in two phases, first it captures a snapshot of the present traffic scenario and then in the second phase it decides on the optimal traffic signal schedule. To grasp the present picture of the traffic conditions, the total number of vehicles in the lane must be determined. Inductive loops \cite{anujsharma2018, gadja2001} which are a very common and relatively inexpensive apparatus can be used to do this. Not only this, an inductive loop can also provide the added benefit of detecting any speed violations by measuring the time a vehicle is positioned on the loop. Such systems are prevalent in the road traffic system across many countries. Another important aspect of the proposed method is the determination of priority for the individual vehicles. This can be implemented using a camera system in association with a vehicle detection algorithm \cite{chauhan2018vehicle,kamkar2016vehicle,ghosh2019adaptive}. The priority of individual vehicles can be assigned to help formulate a snapshot for the current traffic scenario of the intersection.

\section{Conclusion and Future Works}

The main purpose of this research was to propose a feasible model that can reduce the average time of waiting at critical intersections in a densely populated city. A well planned traffic management system that is capable of foreseeing future traffic load can efficiently solve this problem. The model would be highly beneficial, especially for developing countries as most of them have poor traffic management systems. Our proposed method leverages MiniZinc's modeling capabilities to optimize traffic flow at intersections by minimizing average delay of the cars.

Most of the current implementations use fixed or manual methods. These do not actually consider any situation or instances. This results in a very less optimized management and eventually increases the delay time by a big margin. These also lead to complex deadlocks. The model we proposed significantly reduces the time-delays at intersections. The model is technically and economically very feasible which makes it easier for any authority to install it in their system.  

The biggest limitation of our system is the lack of real life testing. The model might behave in a slightly different way when it will be implemented in real life. In future, we would like to implement this model in real life intersections and then incorporate the changes accordingly. The current implementation of the model uses MiniZinc to develop the artificially intelligent approach that manages the intersections, but we plan to develop a new solver dedicated to our proposed model in the future. We see many prospects and scopes of potential development in this research. 

\bibliographystyle{IEEEtran}
\bibliography{my_bib.bib}
\vspace{12pt}

\end{document}